\documentclass[letterpaper, 10 pt, conference]{ieeeconf}  
\IEEEoverridecommandlockouts 
\usepackage{multicol}
\usepackage[bookmarks=true]{hyperref}
\usepackage{amssymb}
\usepackage{amsmath}
\usepackage{graphicx}
\usepackage{graphics}
\usepackage{clrscode3e}
\usepackage{caption}
\usepackage{subcaption}
\usepackage{parcolumns}
\usepackage{pdfcomment}
\usepackage[rgb]{xcolor}
\usepackage{colortbl}

\usepackage[textwidth=1.5in]{todonotes}

\newcommand{\cspace}{{\cal S}}
\newcommand{\rg}{{\sc rg}}
\newcommand{\os}{{\sc cos}}
\newcommand{\ocv}{{\sc ocv}}
\newcommand{\cvar}[1]{V_{#1}}
\newcommand{\eff}[2]{#1.{\it eff}(#2)}
\newcommand{\algName}{{\sc HBF}}

\definecolor{Gray}{gray}{0.85}
\newcolumntype{g}{>{\columncolor{Gray}}c}
\newcolumntype{w}{>{\columncolor{white}}c}

\title{\LARGE \bf Backward-Forward Search for Manipulation Planning}

\author{Caelan Reed Garrett, Tom\'as Lozano-P\'erez, and Leslie Pack Kaelbling
\thanks{Computer Science and Artificial Intelligence Laboratory,
  Massachusetts Institute of Technology, Cambridge, MA 02139 USA
   {\tt\small caelan@mit.edu,  tlp@mit.edu, lpk@mit.edu} \newline
This work was supported in part by the NSF (grant 1420927). Any opinions, findings, and conclusions or recommendations expressed in this material are those of the author(s) and do not necessarily reflect the views of the National Science Foundation. We also gratefully acknowledge support from the ONR (grant N00014-14-1-0486 ), from the AFOSR (grant FA23861014135), and from the ARO (grant W911NF1410433).}        
}

\begin{document}

\maketitle
\thispagestyle{empty}
\pagestyle{empty}

\begin{abstract}
  In this paper we address planning problems in high-dimensional
  hybrid configuration spaces, with a particular focus on manipulation
  planning problems involving many objects.  We present the {\em
    hybrid backward-forward (\algName{})} planning algorithm that uses
  a backward identification of constraints to direct
  the sampling of the infinite action space in a forward search from the
  initial state towards a goal configuration. The resulting
  planner is probabilistically complete and can effectively construct
  long manipulation plans requiring both prehensile and nonprehensile
  actions in cluttered environments.
\end{abstract}

\section{Introduction}

Many of the most important problems for robots require planning in
high-dimensional hybrid spaces that include both the state of the
robot and of other objects and aspects of its environment.  Such
spaces are {\em hybrid} in that they involve a combination of continuous
dimensions (such as the pose of an object, the configuration of a
robot, or the temperature of an oven) and discrete dimensions (such as
which object(s) a robot is holding, or whether a door has been locked).

Hybrid planning problems have been formalized and addressed in the
robotics literature as {\em multi-modal} planning problems and several
important solution algorithms have been
proposed~\cite{Simeon04,Cambon,Hauser,HauserIJRR11,Dogar12,barry2013hierarchical,Srivastava14}.
A critical problem with these algorithms is that they have relatively
weak guidance from the goal:
as the dimensionality of the domain
increases, it becomes increasingly difficult for a forward-search
strategy to sample effectively from the infinite space of possible
actions.

In this paper, we propose the {\em hybrid backward-forward} (\algName)
algorithm.  Most fundamentally, \algName{} is a forward search in
state space, starting at the initial state of the complete domain,
repeatedly selecting a state that has been visited and an action that
is applicable in that state, and computing the resulting state, until
a state satisfying a set of goal constraints is reached.  However, a
significant difficulty is that the branching factor is
infinite---there are generally infinitely many applicable actions in
any given state.  Furthermore, random sampling of these actions will
generally not suffice: actions may need to be selected from very small
or even lower-dimensional subspaces of the space of applicable
actions, making it a measure 0 event to hit an appropriate one at
random.  For example, consider the problem of moving the robot base to
some configuration that may be the initial step of a plan to pick up
and move an object.  That configuration has to be selected from the
small subset of base poses that have an inverse-kinematics solution
that allows it to pick up the object.  Or, in trying to select
the grasp for an object, it may be critical to select one in a smaller
subspace that will allow the object to be subsequently used for a task
or placed in a particular pose.

\algName{} solves the exact planning problem by strongly focusing the
sampling of actions toward the goal.  It uses backward search in a
simplified problem space to generate sets of actions that are {\em
  useful}: these useful actions are components of successful plans in
the simplified domain and might plausibly be contained in a plan to
reach the goal in the actual domain.  The backward search constructs a
{\em reachability graph}, working backward from the goal constraints
and using them to drive sampling of actions that could result in
states that satisfy them.  In order to to make this search process
simpler than solving the overall problem, the constraints in the final
goal, as well as constraints that must hold before an action can be
applied, are considered independently.  Focusing the action sampling
in this way maintains completeness while providing critical guidance
to the search for plans in high-dimensional domains.

\begin{figure}
  \centering
  \includegraphics[width=.31\textwidth]{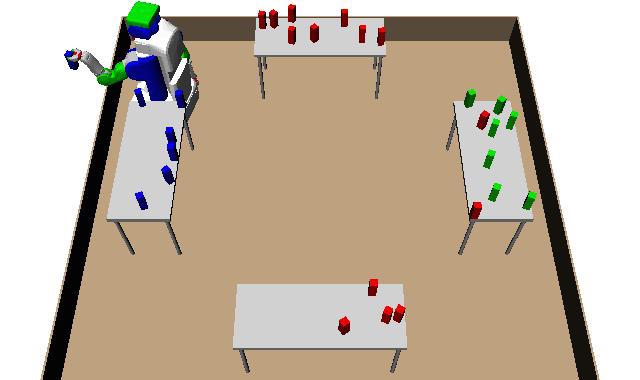}
  \caption{A near-final state of a long-horizon problem.} \label{figure:separate28end}
\end{figure}

\subsection{Related work}

Our work draws from existing approaches to robot 
manipulation planning, integrated task and motion planning, and
symbolic AI planning.

In manipulation planning, the objective is for the robot to operate on
objects in the world.  The first treatments considered a continuous
configuration space of both object placements and robot
configurations, but discrete
grasps~\cite{LozanoPerez81,handeyICRA87,Wilfong89}; they were more
recently extended to selecting from a continuous set of grasps and
formalized in terms of a {\em manipulation graph}~\cite{Alami91,
  AlamiTwoProbs}.  The approach was extended to complex problems
with a single movable object, possibly requiring multiple
regrasps, by using probabilistic roadmaps and a search decomposition
in which a high-level sequence of transit and transfer paths is first
identified, and then motion planning attempts to achieve it~\cite{Simeon04}.

Hauser~\cite{Hauser, HauserIJRR11} identified a generalization of
manipulation planning as {\em multi-modal planning}, that is, planning for
systems with multiple (possibly infinitely many) {\em modes},
representing different constraint sub-manifolds of the configuration
space.  Plans alternate between moving in a single mode, where the
constraints are constant, and switching between modes.  Hauser
provided an algorithmic framework for multi-modal planning that is
probabilistically complete given effective planners for moving within
a mode and samplers for transitioning between modes.  Barry et
al.~\cite{barry2013hierarchical} addressed larger multi-modal planning
problems using a bi-directional RRT-style search, combined with a
hierarchical strategy, similar to that of \algName{}, of suggesting
actions based on simplified versions of the planning problem.

Many recent approaches to manipulation planning
integrate discrete task planning and continuous motion planning
algorithms;  they pre-discretize grasps and placements so that a
discrete task planner can produce candidate high-level plans, then 
use a general-purpose robot motion planner to
verify the feasibility of candidate task plans on a
robot~\cite{dornhege09icaps,dornhege13irosws,Erdem,LagriffoulDSK12}. 
Some other systems combine the task planner and motion planner
more intimately; although they generally also rely on discretization,
the sampling is generally driven by the
task~\cite{Cambon,Plaku,HPN,Pandey12,Srivastava14,GarrettWAFR14}.

Effective domain-independent search guidance has been a major
contribution of research in the artificial intelligence planning
community, which has focused on state-space search methods; they solve
the exact problem, but do so using algorithmic heuristics that quickly
solve approximations of the actual planning task to estimate the
distance to the goal from an arbitrary
state~\cite{BonetG99,bonet2001planning}.  One effective approximation
is the ``delete relaxation'' in which it is assumed that any effect,
once achieved by the planner, can remain true for the duration of the
plan, even if it ought to have been deleted by other
actions~\cite{HoffmannN01}; the length of a plan to achieve the goal
in this relaxed domain is an estimate that is the basis for the
$\proc{H}_{\rm FF}$ heuristic.

The most closely related approach to \algName{} integrates 
symbolic and geometric search into one combined problem and provides
search guidance using an adaptation of the $\proc{H}_{\rm FF}$ heuristic to
directly include geometric considerations~\cite{GarrettWAFR14}.  It
was able to solve larger pick-and-place problems than most previous
approaches but suffered from the need to pre-sample its geometric
roadmaps. 

\subsection{Problem formulation}

Our objective is to find a feasible plan in a hybrid configuration
space.
A {\em configuration variable} $\cvar{i}$ defines some aspect or
  dimension of the overall system; its values may be drawn from a
  discrete or continuous set.
A {\em configuration space} $\cspace$ is the Cartesian product of the
  domains of a set of configuration variables $\cvar{1}, \ldots,
  \cvar{n}$.
A {\em state} $s = \langle v_1, \ldots, v_n \rangle \in \cspace$
  is an element of the configuration 
  space consisting of assignments of values to all of its
  configuration variables.

A {\em simple constraint} $C_i$ is a restriction on the possible values
  of state variable $i$: it can be an equality $V_i = C_i$, or a set
  constraint $V_i \in C_i$, where $C_i$ is a subset of the domain of
  $V_i$.  
A {\em constraint} $C_{i_1, \ldots, i_k}$ is a restriction on
  the possible values of a set of state variables $V_{i_1}, \ldots,
  V_{i_k}$,  constraining $(V_{i_1}, \ldots, V_{i_k}) \in C_{i_1,
    \ldots, i_k}$, where $C_{i_1, \ldots, i_k}$ is a subset of the
  Cartesian products of the domains of $V_{i_1}, \ldots,
  V_{i_k}$.  We will assume that there is a finite set of {\em
    constraint types} $\xi \in \Xi$; each $\xi$ specifies a
  constraint using a functional form so that $C_{i_1, \ldots, i_k}$
  holds whenever $\xi(V_{i_1}, \ldots, V_{i_k}) = 0$.

A state $s = \langle v_1, \ldots, v_n\rangle$ {\em satisfies} a 
  constraint $C_{i_1, \ldots, i_k}$ if and only if $\langle v_{i_1}, \ldots,
  v_{i_k} \rangle \in C_{i_1, \ldots, i_k}$.  
An {\em effect} $E_i$ is an equality constraint on $V_i$, used
  to model the effect of taking an action on that variable.

An {\em action} $a$ is an effector primitive that may be
  executed by the robot.  It is characterized by a set of {\em condition
  constraints} $C^1, \ldots, C^K$ on some subset of the the
  configuration variables (written $a.con$), and a set of effects
  $E^1, \ldots, E^l$ on 
  some generally distinct set of configuration variables.  
    For any state
    $s = \langle v_1, \ldots, v_n\rangle$ that satisfies constraints
    $C^1, \ldots, C^K$, in which the robot executes action $a$, the
    resulting state $\eff{a}{s}$ has $V_i = v_i$ for all configuration
    variables not mentioned in the effects, and value
    $V_{I(E^j)} = E^j$ for all configuration variables in the effects
    set.  (We use $I(E)$ to denote the index of some particular
    effect).  
A {\em planning problem} is an initial state $s_0 \in \cspace$
  and goal set of constraints $\Gamma$. 
A {\em plan} is a sequence $a_1, \ldots, a_m$ of actions;  it
  {\em solves} a planning problem $s_0, \Gamma$ if
  $\eff{a_m}{\eff{\ldots a_1}{s_0}} \in \Gamma$. 

\section{HBF algorithm}
In this section, we describe the ~\algName{} algorithm in detail,
beginning with the top-level forward search, continuing with the
definition of a reachability graph, then showing how focused action sampling
and a heuristic can be derived from the reachability graph.

\subsection{Forward search: persistent enforced hill-climbing}

Figure~\ref{forwardAlg} shows a basic framework for persistent
enforced hill-climbing; we use this form of search because it can
maintain completeness with infinite action spaces and because attempts
to find a path of optimal length can quickly become bogged down with a
large agenda.  It is given as input an initial state $s_0$, a set of
goal constraints $\Gamma$, and a set of possible actions ${\cal A}$.
It keeps a queue of search nodes, each of which stores a state
$s \in \cspace$, an action $a \in {\cal A}$, and a pointer to its
parent node in the tree.  A heuristic function $\proc{h}$, which we
discuss in section~\ref{heuristic}, is used to compute an estimate of
the distance to reach a state satisfying $\Gamma$ from state $s$.  The
search enforces hill-climbing, by remembering the minimum heuristic
value $h_{\rm min}$ seen so far.  Whenever a state is reached with a
lower heuristic value, the entire queue is popped, leaving only the
initial state (in case this branch of the tree is a dead-end) and the
current state on the queue.  If the queue has not been reset, it adds
the node that it just popped off back to the end of the queue.  This
last step is critical for achieving completeness in domains with an
infinite action space: it ensures that we consider taking other
actions in the state associated with that node.

Intuitively, then, whenever a state with an improved heuristic value
is reached, the queue is left containing just $s_0$ and the most
recent state (call it $s$).  Then, until a better state is reached, we
will: sample a child $s^1_0$ of $s_0$ (now the queue is: $s$, $s^1_0$,
$s_0$); sample a child $s^1$ of $s$ (now the queue is: $s^1_0$,
$s_0$, $s^1$, $s$), sample a child $s^{(1, 1)}_0$ of $s^1_0$ (now
the queue is $s_0$, $s^1$, $s$, $s^{(1, 1)}_0$, $s^1_0$), sample a new
child $s^2_0$ of $s_0$ (now the queue is  $s^1$, $s$, $s^{(1,
  1)}_0$, $s^1_0$, $s^2_0$, $s_0$), and so on.  This
search strategy is as driven by the heuristic as possible, but takes
care to maintain completeness through the ``persistent'' sampling at
different levels of the search tree.

We compactly characterize the set of
possible actions using action templates with the following form:\\
\noindent
\proc{ActionTemplate}$(\theta_1, \ldots, \theta_k)$:\\
$\begin{array}{ll}
\id{con:} & (q_{i_1}, \ldots, q_{i_t}) \in C_{i_1, \ldots, i_t}(\theta) \\
& ... \\
\id{eff:} & q_j = e_j(\theta)\\
& ... \\
\end{array}$ \\

An action template specifies a generally infinite set of actions, one
for each possible binding of the parameters $\theta_i$, which may be
drawn from the domains of state variables or other values such as
object designators.  We assume that action templates are never
instantiated with a set of parameters that violate permanent
constraints (e.g., exceeding joint limits or colliding with immovable objects
objects). 

In addition to the heuristic, we have also left indeterminate how an
action or set of actions is chosen to be applied to a newly reached
state $s$.  To do so, we must introduce the notion of a reachability
graph. 

\begin{figure}
\begin{codebox}
\Procname{$\proc{forward-search}(s_0, \Gamma, {\cal A})$}
\li $h_{\rm min} = \proc{h}(s_0, \Gamma)$
\li $Q = \proc{Queue}(\proc{Node}(s_0, \kw{None}, \kw{None}))$
\li \While \kw{not} $\proc{empty}(Q)$: \Do
\li $n = \proc{pop}(Q)$; $reset =  \kw{False}$
\li \kw{if} $n.s \in \Gamma$: \kw{return} $\proc{retrace-plan}(n)$
\li \For $a \in \proc{sample-actions}(s)$ \If \kw{not} \id{reset}: \Do \label{app}
\li $s' = a.\id{eff}(n.s)$; $h' = \proc{h}(s', \Gamma)$
\li \If $h'< h_{\rm min}$ \Then
\li $h_{\rm min} = h'$; \id{reset} = \kw{True}
\li $Q = \proc{Queue}(\proc{Node}(s_0, \kw{None}, \kw{None}))$
\End
\li $\proc{push}(Q, \proc{Node}(s', a, n))$
\End
\li \If \kw{not} $reset$: $\proc{push}(Q, n)$ \label{addN}
\End 
\li \kw{return} \kw{None}
\End
\end{codebox}
\caption{Top-level forward search.}
\label{forwardAlg}
\end{figure}

\subsection{Reachability graphs}

A {\em reachability graph} (\rg) is an oriented hyper-graph in which:
(1) each vertex is a point in (a possibly lower-dimensional subspace
of) the configuration space $\cspace$, which can be represented as an
assignment of values to a set of configuration variables $\{V_{i_1} =
  v_{i_1}, \ldots, V_{i_k} = v_{i_k}\}$; 
(2) each edge is labeled with an action $a$;
(3) the {\em outgoing} vertices of the edge are the effects of $a$; and
(4) the {\em incoming} vertices of the edge are assignments that
  collectively satisfy the condition constraints of $a$. 

We say that an \rg{} $G$ contains a {\em derivation} of constraint
$C$ from initial state $s$ if either (1) $C$ is satisfied in $s$ or
(2)~there exists an action $a \in G$ that has $C$ as an effect and
for all $C_i \in A.\id{con}$, $G$ contains a
derivation of $C_i$ from $s$.  

To construct the \rg{}, we solve a version of the planning problem
that is simplified in multiple ways, allowing very efficient
identification of useful actions.  There are three significant
points of leverage: (1) individual constraints in the goal and in the
conditions for an action are solved independently; (2) 
it is possible to plan efficiently in low-dimensional subspaces (for example,
by using an {\sc rrt} to find a path for the robot alone, or for an
object alone); and (3) plans made in low-dimensional subspaces are
allowed to violate constraints from the full space (for example,
colliding with an object) and the system subsequently plans to achieve
those constraints that were violated (for example, by moving the
object that cause the collision out of the way).

The fundamental organizing idea is the {\em constrained operating 
  subspace} (\os).  A \os{} is a subspace of \cspace{} defined by two
restrictions: first, it is restricted to a set of {\em operating
  constraint variables} (\ocv's), and then further restricted to the
intersection of that space with a set of constraints ${\cal C}$ that are
defined in terms of the \ocv's.  A \os{} 
provides methods, used in planning, for generating potential samples in that
subspace and also for generating action instances to move into and
within the subspace.  Every \os{} $\Omega$ is defined by:
\begin{itemize}
\item A set ${\cal V} = \{V_{i_1}, \ldots, V_{i_k}\}$ of configuration variables;
\item A set ${\cal C}$ of constraints;
\item An {\em intersection sampler}, which generates samples from
  ${\cal C} \cap C_i$ for some domain constraint $C_i$ of one of the
  functional forms in $\Xi$;
\item A {\em transition sampler}, which generates samples $(o, a, o')$, 
  where $o' = \langle v_{i_1}, \ldots, v_{i_k}\rangle \in \Omega$,
  $a$ is an action whose effects include (but need not be limited to)
  $V_{i_1} = v_{i_1}, \ldots, V_{i_k} = v_{i_k}$, and $o$ is an
  element of a different \os{}.
\item  A {\em roadmap sampler}, which generates samples
  $(o, a, o')$  where $o \in \Omega$, $o' \in \Omega$, and $a$ is an
  action such that
  the initial abstract state $o$ satisfies the condition constraints
  of $a$ that apply to $\Omega$'s variables and the effects of $a$
  include $o'$.
\end{itemize} 
Planning within a \os{} is completely restricted to that subspace,
observing only constraints involving values of variables in ${\cal
  V}$. 
Our definition of a \os{} is in contrast to 
Hauser's definition of a mode, in that Hauser's modes are defined
on the complete configuration space, so that every variable either is
fixed or can be controlled by the system.

In a domain with infinite state and action spaces, the corresponding
\rg{} will generally be infinite as well.  We will iteratively
construct a subgraph of the full \rg{} that can derive a set of goal
constraints $\Gamma$ from a state $s$. The \rg{} subgraph will
eventually contain actions that are the first step of a solution to
the planning problem $(s, \Gamma)$.  Because it considers the
satisfaction of constraints independently, it may be that there are
actions that produce all of the condition constraints of some other
action, but those actions are incompatible in a way that implies that
this collection of actions could not, in fact, be sequenced to provide
a solution. For this reason, during the forward search, we will return
to the process of growing the \rg{}, until it generates actions that
are, in fact, on a feasible plan.
  
Procedure \proc{ReachabilityGraph} creates a new \rg{}, with start state
$s$ as a vertex (meaning that the assignment of values to variables in
$s$ can be used to satisfy conditions of edges that are added to the
\rg{}) and adds an agenda item for each constraint in the goal; these
constraints are associated with the dummy action $\id{goalSat}$, which,
when executable implies that the overall relaxed goal is satisfied.

Procedure \proc{GrowRG} takes as input a partial \rg{} $G$ (either just
initialized or already previously partially grown), a set of \os's
$\Sigma$, and a time-out parameter $\tau$.  We assume that it, as well
as all of the \os{} samplers in $\Sigma$, have a set of action
templates ${\cal A}$ available to them.  It grows the \rg{} for $\tau$
steps by popping a $C, a$ pair off of the queue, and trying to find
actions that can help achieve $C$ and adding them to the \rg{}.  It
begins by generating sample elements $o$ of the intersection between
$\Omega$ and $C$ using \proc{s-intersection}.  It adds them to the set of vertices and connects
them as ``witnesses'' that can be used to satisfy the $C$ condition of
action $a$.   Now the transition sampler (\proc{s-transition}) seeks $(o, a', o')$ triples, where
$o \in \Omega$, $o' \in \Omega'$,  and $o$ reaches $o'$ via
action $a'$.  Finally, \proc{s-roadmap} samples $(o, a', o')$ triples
and adds vertices and hyperedges
that move within $\Omega$.  
Each condition of each new edge is added
to the queue, and the original node is placed back in the queue for
future expansion.

\begin{figure}
\begin{codebox}
\Procname{$\proc{ReachabilityGraph}(s, \Gamma):$}
\li $V, E = \{s\}, \{\:\}$
\li $Q = \proc{Queue}()$
\li \For $C \in \Gamma$: $\proc{push}(Q, (C, \id{goalSat}))$ \label{init}
\end{codebox}
\begin{codebox}
\Procname{$\proc{grow-rg}(G, \Sigma, \tau):$}
\li $V, E = G.V, G.E$; $Q = V.Q$; $t=0$
\li \While \kw{not} $\proc{empty}(Q)$ \kw{and} $t < \tau$: \Do
\li $C, a = \proc{pop}(Q)$; $t$++
\li \For $\Omega \in \{\Omega \in \Sigma \mid \Omega \cap C \neq
\emptyset \}$: \Do 
\li \For $o \in \proc{s-intersection}(\Omega, C):$ \Do
\li $\proc{connect}(o, C, a)$; $V \:\cup\!= \{o\}$ \label{connect}
\End
\li $V', E' = \proc{s-transition}(\Omega, V)$
\li $V', E' \:\cup\!= \proc{s-roadmap}(\Omega, V \cup V')$ \label{moveModes}
\li \For $a' \in E':$ \Do
\li \For $C' \in a'.\id{con}$: \Do
\li $\proc{push}(Q, (C', a'))$
\End\End
\li $V, E \:\cup\!= V', E'$
\End
\li $\proc{push}(Q, (C, a))$
\End
\End
\end{codebox}
\caption{Construction and growth of the reachability graph.}
\label{backwardAlg}
\end{figure}

\subsection{Action sampling and heuristic}
\label{heuristic}

Now, we can revisit the forward search algorithm in
figure~\ref{forwardAlg} to illustrate the points of contact with the
backward algorithm.  
The \rg{} for $s$ is first used to compute its heuristic value.   The
heuristic procedure \proc{H} initializes an \rg{} for $(s, \Gamma)$ if
necessary;  if the \rg{}
does not yet contain a derivation for all the constraints in $\Gamma$
from $s$, then the graph is grown in an attempt to find such a
derivation. A heuristic estimate of the distance to reach $\Gamma$
from $s$ is the number of actions in a derivation. Because computing the
minimum length derivation is NP-hard, we use the $\proc{H}_{\rm FF}$ algorithm
which greedily chooses a small derivation~\cite{HoffmannN01}. Because the \rg{}
and {\em relaxed plan graph} of the $\proc{H}_{\rm FF}$ algorithm both make a similar
independence approximation, $\proc{H}_{\rm FF}$ can be computed on top of the \rg{}
with minimal algorithmic changes. 

The \proc{sampleActions} procedure seeks actions from the \rg{} that are applicable
in $s$ but have not yet been applied.  If there are no such actions,
then the \rg{} is grown for $\tau$ steps.  In our experience, it is
frequently the case that as soon as a derivation is found,
there is at least one action in the \rg{} that is feasible for use in
forward search.  To increase performance in practice, the first set of actions returned are
restricted to the {\em helpful actions} computed by the $\proc{H}_{\rm FF}$ algorithm~\cite{HoffmannN01}.

\algName{} can be shown to be probabilistically complete if the \os{}
samplers are themselves probabilistically complete. The intuition
behind the argument is that the reachability graph constructs a
superset of the set of solutions, due to its independence
approximation.  The proof is similar in nature to the probabilistic
completeness proof by Hauser~\cite{HauserIJRR11}.  Due to space
limitations, a sketch of this argument can be found at the following
URL: \url{http://web.mit.edu/caelan/www/research/hbf/}.

\section{Domain and example}

In this section, we describe aspects of the domain formalization used
in the experiments of the last section and work through an
illustrative example.  \algName{} is a general-purpose planning
framework for hybrid domains; here we are describing a particular
instance of it for mobile manipulation.  

Our domain includes a mobile-manipulation robot (based on the Willow
Garage PR2) that can pick up, place, and push small rigid objects.
The configuration spaces for problem instances in this domain are
defined by the following configuration variables: $r$ is the robot's
configuration, $o_i$ is the pose of object $i$, $h$ is a discrete
variable representing which object the robot is holding (its value is
the object's index or \proc{none}), and $g$ is the {\em grasp
  transform} between the held object and the gripper or \proc{none}.

\subsection{Action templates}

The robot may move from configuration $q$ to configuration $q'$ if it
is not holding any object and there is a collision-free trajectory,
$\tau$, from $q$ to $q'$ given the poses of all the objects. Each
object $i$ also has a constraint that its pose $o_i$ be in the set of
poses that do not collide with $\tau$. This set is represented
by $\proc{c-free-poses}_i$.

\noindent \proc{Move}$(q, q', \tau)$:\\
$\begin{array}{ll}
\id{con:} & r, h = q, \kw{None} \\
& o_i \in \proc{c-free-poses}_i(\tau) \:\forall i \\
\id{eff:} & r = q' \\
\end{array}$ \\

If the robot is holding object $j$ in grasp $\gamma$, it may move from
configuration $q$ to $q'$; the free-path constraint is extended to
include the held object and the effects are extended to include the
fact that the pose of the held object changes as the robot moves.  The
values of $r$ and $o_i$ change in such a way that $g$ is held
constant.

\noindent \proc{MoveHolding}$(q, q', \tau, j, \gamma)$:\\
$\begin{array}{ll}
\id{con:} & r, h, g = q, j, \gamma \\
& o_i \in \proc{c-free-poses-holding}_i(\tau, j, \gamma) \:\forall i \neq j\\
\id{eff:} & r, o_j = q', \proc{pose}(q', \gamma) \\
\end{array}$ \\

The \proc{Pick} action template characterizes the state change between
the robot holding nothing and the robot holding (being rigidly
attached to) an object.  It is parameterized by a robot configuration
$q$, an object to be picked $j$, and a grasp transformation $\gamma$.
If the robot is in the appropriate configuration, the hand is empty,
and the object is at the pose obtained by applying transformation
$\gamma$ to the robot's end-effector pose when it is in configuration
$q$, then the grasp action will succeed, resulting in object $j$ being
held in grasp $\gamma$.  No other variables are changed. The \proc{Place}
template is the reverse of this action. \proc{Unstack} and
\proc{Stack} action templates can be defined similarly by augmenting
\proc{Pick} and \proc{Place} with a parameter and constraint involving
the base object.

\begin{parcolumns}[colwidths={1=1.5 in},nofirstindent]{2}
\colchunk{\proc{Pick}$(q, j, \gamma)$:\\
$\begin{array}{ll}
\id{con:} & r, h = q, \kw{None} \\
& o_j = \proc{Pose}(q, \gamma) \\
\id{eff:} & h, g = j, \gamma \\
\end{array}$}

\colchunk{\proc{Place}$(q, j, \gamma)$:\\
$\begin{array}{ll}
\id{con:} & r, h, g = q, j, \gamma \\
\id{eff:} & h = \kw{None} \\
& o_j = \proc{Pose}(q, \gamma) \\
\end{array}$ \\}\colplacechunks
\end{parcolumns}

The \proc{Push} action is similar to \proc{MoveHolding} except that
grasp transformation parameter $g$ is replaced with an initial pose
parameter $p$ for the object.  Procedure $\proc{end-pose}_j$ computes
the resulting pose of the object if the robot moves from $q$ to $q'$
along a straight-line path. \\
\noindent \proc{Push}$(q, q', j, p)$:\\
$\begin{array}{ll}
\id{con:} & r, h, o_j = q, \kw{None}, p \\
& o_i \in \proc{c-free-poses-push}_i(q, q', j, p) \:\forall i \neq j\\
\id{eff:} & r, o_j = q', \proc{end-pose}_j(q, q', p) \\
\end{array}$ \\

\subsection{Constrained operating subspaces}

Action templates define the fundamental dynamics of the domain;  but
to make \algName{} effective, we must also define the \os{}s that
will be used to construct \rg{}s, selection actions, and compute the
heuristic.  
\begin{itemize}
\item $\Omega_\id{move}$ considers the robot configuration, ${\cal V}
  = \{r\}$, and has no constraints.
\item $\Omega_{\id{moveHolding}(i, g)}$ is parameterized by an object
  $i$ and grasp $g$;  it considers the robot configuration and pose of
  object $I$, ${\cal V} =   \{r, o_i\}$ subject to the grasp
  constraint ${\cal C} = \{o_i =   \proc{pose}(r, g)\}$.
\item $\Omega_{\id{push}(i)}$ is parameterized by an object $i$; it
  abstracts away from the robot configuration and considers only the
  object's pose, ${\cal V} = \{o_i\}$, subject to the constraint that
  $o_i \in \id{table}$ (where \id{table} is the space of poses for
  $o_i$ for which it is on the table.)
\item $\Omega_{\id{place}(i)}$ is parameterized by object $i$ for graspable objects; it also
    considers only the object's pose, ${\cal V} = \{o_i\}$, subject to
    the constraint ${\cal C} = \{o_i \in \id{table}\}$.
\item $\Omega_{\id{pick}(i)}$ has the same parameters, ${\cal V}$ and
    ${\cal C}$ as $\Omega_{\id{place}(i)}$.
\end{itemize}

Each of these \os{}'s is augmented with sampling methods.
Placements on flat surfaces and other objects, as well as
placements in regions, are generated using Monte Carlo rejection
sampling. Inverse reachability and inverse kinematics sample robot
configurations for manipulator transforms. The \proc{s-roadmap}
samplers for moving and moving while holding are bidirectional RRT's
that plan for the robot base then robot arm. The \proc{s-roadmap}
sampler for pushing is also a bidirectional RRT in the configuration
space of just an object. Each edge of the tree is an individual push action
computed by a constrained workspace trajectory planner that samples a
robot trajectory that can perform the push.

\subsection{Example}

\begin{figure}
  \centering
  \includegraphics[width=0.3\textwidth]{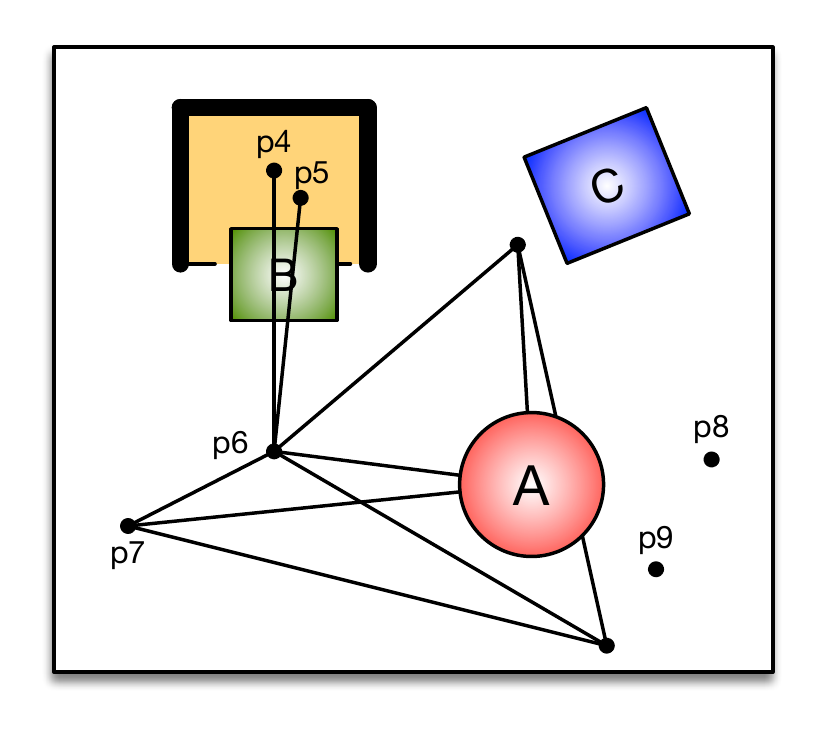}
  \caption{An example initial state, with roadmap for object $A$.}
\label{example}
\end{figure}

Figure~\ref{example} shows a top-down view of a table with three
movable objects and a fixed u-shaped obstacle.  This is the initial
state $s_0$ of a planning problem for which the goal $\Gamma =
\{o_A \in R \}$, where $R$ is the region of configuration space for
object $A$ that would place it on the table within the
  fixed obstacle.  Assuming that the robot is in some initial
  configuration $q_0$, we can describe $s_0$ as
$\langle r=q_0, o_A= p_1, o_B = p_2, o_C = p_3, h = \id{none}, g =
\id{none} \rangle$.  Objects $B$ and $C$ can be grasped but $A$ cannot.

\begin{figure*}
  \centering
  \includegraphics[width=\textwidth]{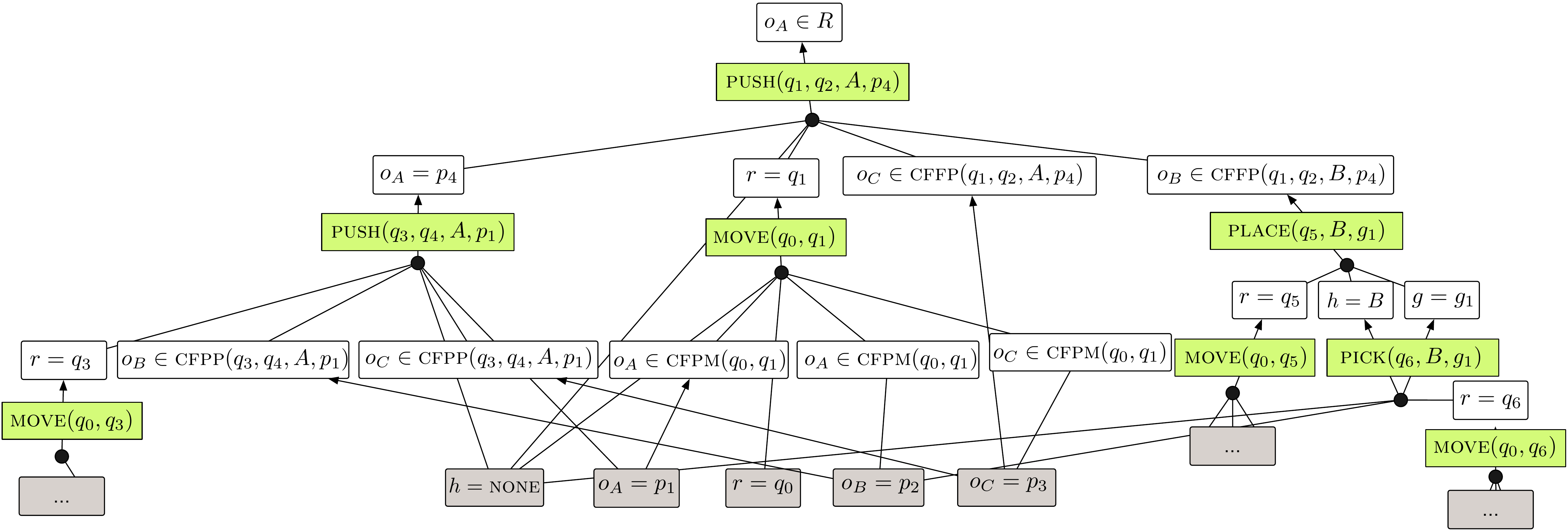}
  \caption{Building the reachability graph.}
\label{rg}
\end{figure*}

To begin the forward search, \algName{} must construct an initial
\rg{}, which will be used to compute the initial heuristic value
$h_\id{min}$ and to generate the first set of actions in the search.
In the following, we will sketch the process by which the \rg{}, $G$,  is
constructed, showing parts of $G$ in figure~\ref{rg}.
We add $s_0$ to the set of vertices, add a dummy action with $\Gamma$
as its conditions constraints to the set of edges, and 
initialize $Q=[(o_A \in  R, \id{goalSat})]$.  

{\bf Iteration 1:} $C,a = (o_A \in  R, \id{goalSat})$ is popped
from $Q$.  The only $\Omega$ that has a non-null intersection with $C$ is
$\Omega_{\id{push}(A)}$, so ${\cal V} = \{o_A\}$ and
${\cal C} = \{o_A \in \id{table}\}$.  The first step is to
generate a set of samples of $o_A$ in
$R \cap \id{table} = R$;
poses $o_A = p_4$ and $o_A = p_5$ are such samples (note that we
automatically reject sampled poses that would cause a collision with a
permanent object), which we add to $G$.  Next, we would use
\proc{s-transition} to generate samples $(o, a, o')$; however,
because object $A$ cannot be grasped, there are no other \os{}s that
can be connected to $\Omega$.  Finally, we use \proc{s-roadmap} to
generate samples $(o, a, o')$ that move within
$\Omega_{\id{push}(A)}$.  This generates several more vertices, shown
as $o_A = p_6$ and $o_A = p_7$.  The values of $a$ are complete ground
\proc{push} action instances; we show two particular ones (that
connect up to the initial pose of $A$, $p_1$) in the figure.  We add
the condition constraints of the new actions that cannot be satisfied
using an existing vertex to the agenda, as well as putting the
original item back on.  At this point, the agenda is:
$[(o_B \in \proc{cfpp}(q_1, q_2, A, p_4),\proc{push}(q_1, q_2, A,
p_4)),\\ (r = q_1, \proc{push}(q_1, q_2, A, p_4)),\\ (r = q_3,
\proc{push}(q_3, q_4, A, p_1)), \\(o_A \in R, \id{goalSat})]$.
(\proc{cfpp} is an abbreviation for
\proc{c-free-poses-push}.)   The first constraint is that object $B$
not be in the way of the final push of $A$;  the second two are that
the robot be in the necessary configuration to perform each push.

{\bf Iteration 2:} From $Q$, we pop \\
$C, a = (o_B \in \proc{cfpp}(q_1, q_2, A, p_4), \proc{push}(q_1, q_2,
A, p_4))$.
There are multiple \os{}s that have non-null intersection with $C$,
and they would each be used to generate new vertices and edges for
$G$.  In our exposition, we will focus on $\Omega_{\id{place}(B)}$, so
that ${\cal V} = \{o_B\}$ and ${\cal C} = \{o_B \in \id{table}\}$.
The intersection of constraints is that
$o_B \in \proc{cfpp}(q_1, q_2, A, p_4)$, and so we can generate
samples of poses for $B$ that are not in the way of pushing $A$.
Vertices like $o_B = p_8$ and $o_B = p_9$ are added to $G$.  Next, we
use \proc{s-transition} to generate samples $(o, a, o')$ that connect
from other \os{}s; in particular, to move in from
$\Omega_{\id{pick}(B)}$ we might sample action
$\proc{place}(q_5, B, g_1)$ where $p_8 = \proc{pose}(q_5, g_1)$.
Finally, \proc{s-roadmap} is unable to generate samples that move
within $\Omega_{\id{place}(B)}$.  At this point, we can add the
\proc{place} action to $G$, as well as its conditions, and add its
unmet conditions to the agenda.  The agenda will be
augmented with
$((r = q_5, \proc{place}(q_5, B, g_1)), (h = B, \proc{place}(q_5, B,
g_1)), (g = g_1, \proc{place}(q_5, B, g_1))$.

{\bf Subsequent iterations:}  In subsequent iterations, an instance of
\proc{pick} for object $B$ will be added, which can satisfy both the
$h = B$ and $g = g_1$ conditions of the \proc{place} action, and
several \proc{move} actions will need to be added to satisfy the robot
configuration conditions.  Figure~\ref{rg} shows a subgraph of $G$
containing one complete derivation of $\Gamma$.  Nodes in gray with
labels in them are vertices that are part of $s_0$.  Three \proc{move}
actions are shown with their condition constraints elided for space
reasons, but they will be analogous to the one \proc{move} action that
does have its conditions shown in detail.  The green labels show the
actions that are part of this derivation;  there are 8 of them, which
would yield an $\proc{H}_{\rm FF}$ value of 8 for $s_0$, unless there was a
shorter derivation of $\Gamma$ also contained in $G$.  But, in fact,
the shortest plan for solving this problem involves 8 steps, so the
heuristic is tight in this case.  Additionally, all of the actions in
$G$ whose conditions are satisfied in $s_0$ are applied by the forward
search;  so all four \proc{move} actions will be considered in the
first iteration.

This example illustrates the power that the \rg{} has in guiding the
forward search;  the cost of computing it is non-negligible, but it is
able to leverage searches in low-dimensional spaces using
state-of-the-art motion-planning algorithms very effectively.  Two
very important things happened in the process illustrated above.
First, in planning how to push $A$ into the goal region, the system
was able to first reason only about the object and later worry about
how to get the robot into position to perform the pushing actions.
Second, it was able to plan the pushes without worrying about the fact
that object $B$ was in the way, and then later reason about how to
clear the path.  This style of backward goal-directed reasoning is
very powerful and has been used effectively to solve {\sc
  namo} (navigation among movable obstacle) problems~\cite{StilmanWAFR06}
and by the {\sc hpn} planner~\cite{HPN}.


\begin{figure*}
  \centering
  \begin{subfigure}[b]{\textwidth}
    \centering
      \begin{subfigure}[b]{.31\textwidth}
      	\includegraphics[width=\textwidth]{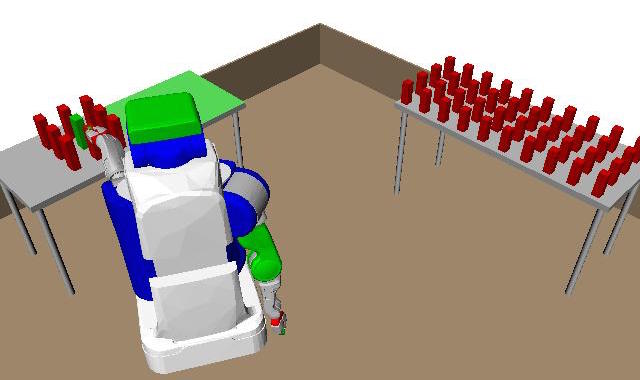} 
	\caption{Problem 1} \label{world:distractions}
      \end{subfigure}
      \begin{subfigure}[b]{.31\textwidth}
      	\includegraphics[width=\textwidth]{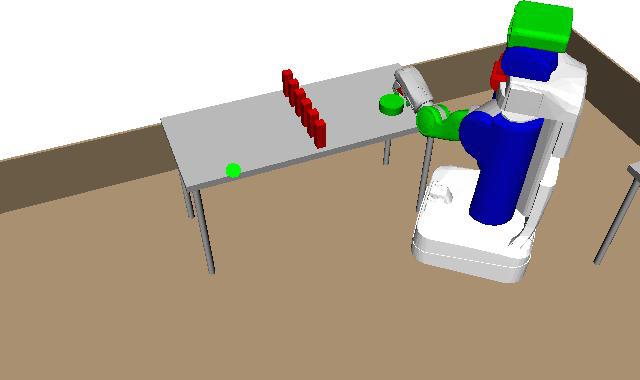} 
	\caption{Problem 2} \label{world:push}
      \end{subfigure}
       \begin{subfigure}[b]{.31\textwidth}
    	\includegraphics[width=\textwidth]{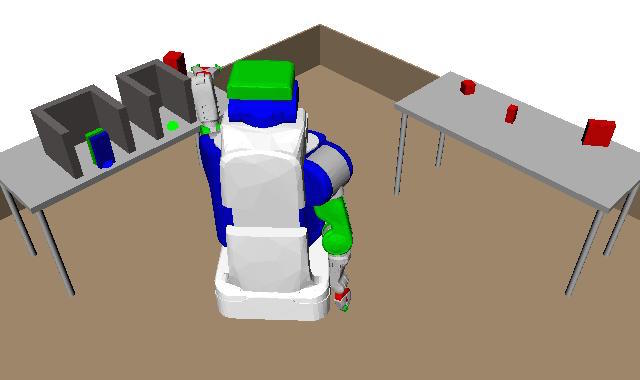}
	\caption{Problem 3} \label{world:regrasp}
      \end{subfigure}   
  \end{subfigure}\\
  \begin{subfigure}[b]{\textwidth}
    \centering
      \begin{subfigure}[b]{.31\textwidth}
    	\includegraphics[width=\textwidth]{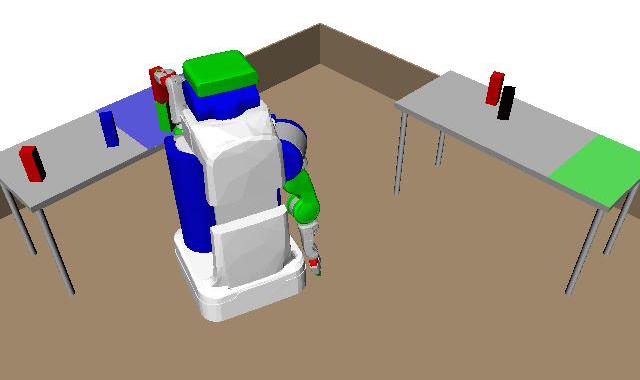}
	\caption{Problem 4} \label{world:stack}
      \end{subfigure}
      \begin{subfigure}[b]{.31\textwidth}
    	\includegraphics[width=\textwidth]{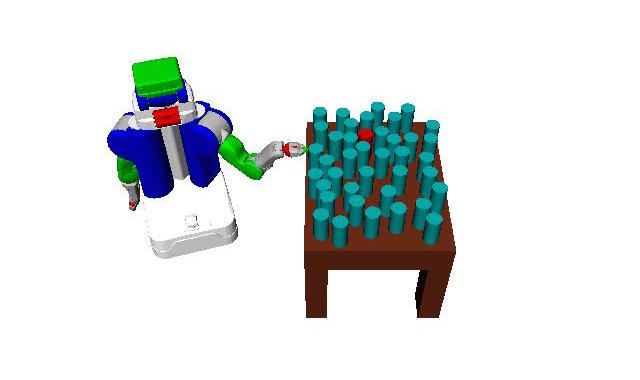}
	\caption{Problem 5} \label{world:srivastava}
      \end{subfigure}
      \begin{subfigure}[b]{.31\textwidth}
    	\includegraphics[width=\textwidth]{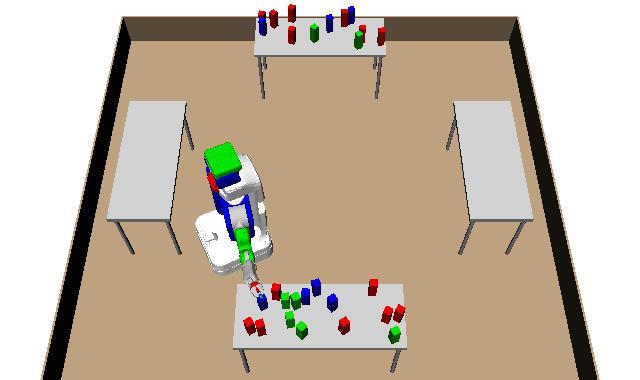}
	\caption{Problem 6} \label{world:separate}
      \end{subfigure}
  \end{subfigure}\\
\caption{An early state on a valid plan for each problem.}
\label{results}
\end{figure*}

\begin{figure*}[t]
\minipage{.71\textwidth}
\begin{tabular}{||c||c|g|c|c||c|g|c|c||}
\hline
P & \multicolumn{4}{|c||}{$\proc{H}_{\rm 0}$}&\multicolumn{4}{|c||}{$\proc{H}_{\rm  FF}$}\\
\hline
&
 \% & runtime & length & visited &
 \% & runtime & length & visited \\
\hline
1 & 
100 & 12 (7) & 8 (0) & 312 (280) &
100 & 4 (1) & 12 (2) & 12 (4) \\
\hline
2 & 
97 & 62 (26) & 16 (0) & 416 (74) &
100 & 7 (1) & 16 (0) & 20 (2) \\
\hline
3 & 
62 & 238 (62) & 16 (0) & 4630 (1424) &
100 & 6 (1) & 16 (0) & 74 (4) \\
\hline
4 & 
0 & 300 (0) & - (-) & 2586 (186) &
97 & 12 (4) & 24 (4) & 112 (48) \\
\hline
5 & 
0 & 300 (0) & - (-) & 3182 (762) &
98 & 23 (9) & 24 (4) & 170 (74) \\
\hline
6 & 
0 & 300 (0) & - (-) & 1274 (80) &
100 & 82 (13) & 72 (4) & 382 (74) \\
\hline
\end{tabular}
\caption{Manipulation experiment results over 60 trials.}
\label{table}
\endminipage\hfill
\minipage{.28\textwidth}
  \includegraphics[width=\textwidth]{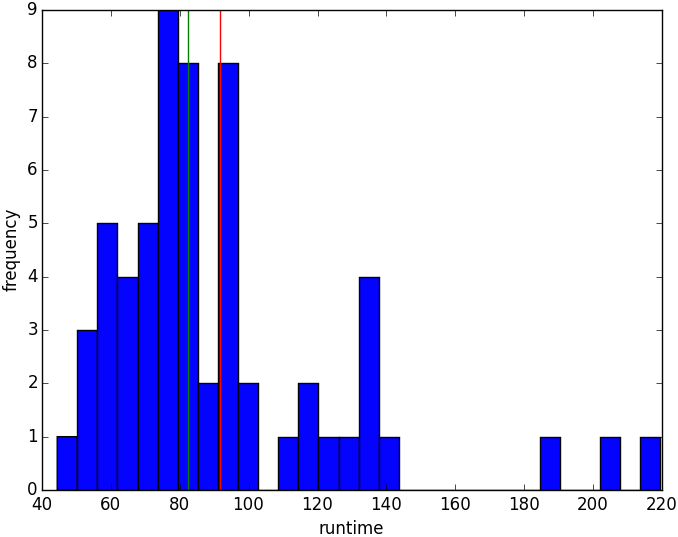}
  \caption{Problem 6 $\proc{H}_{\rm FF}$ runtimes.} \label{figure:histogram}
\endminipage\hfill
\end{figure*}


\section{EXPERIMENTAL RESULTS} 

We applied \algName{} to six different manipulation problems to
characterize its performance.  The planner, samplers, and PR2 robot
manipulation simulations were written in Python using
OpenRAVE~\cite{openrave} without substantial performance optimization.
In each problem, red objects represent movable objects that have no
specified goal constraints.  However, they impose geometric
constraints on the problem and must usually be manipulated in
order to produce a satisfying plan.

\subsubsection*{\bf Problem 1} The goal constraint is for the green
block, which is surrounded by 8 red blocks, to be on the green region
on the table. Notice that the right table has 40 movable red objects
on it that do not block the path of the green object. 

\subsubsection*{\bf Problem 2} The goal constraint is for the green
cylinder to be at the green point on the edge of the table. The
cylinder is too big for the robot to to grasp, so it must push it
instead.  The robot must move several of the red objects and then push
the green cylinder several times to solve the problem.

\subsubsection*{\bf Problem 3} A thin but wide green block starts
behind a similar blue block. The goal constraints are that the green
block be at the green point and that blue block be at the blue
point, which is, in fact, its initial location.  The is problem is
non-monotonic in that the robot must first violate one of the goal
constraints, and then re-achieve it.  Additionally, the right cubby
housing the green goal point is thinner than the left cubby, so the
green block can only be placed using a subset of its grasps, all of
which are infeasible for picking it up at its initial location. This
forces the robot to place and regrasp the green block.  

\subsubsection*{\bf Problem 4} The goal constraints are that the green
block be on the green region of the table, the blue block be on the
blue region of the table, and the black block be on top of the blue
block. Because the black block must end on the blue block, which
itself must be moved, no static pre-sampling of object poses would
suffice to solve the problem.  Additionally, a red block starts on top
of the green block, preventing immediate movement of the green
block. 

\subsubsection*{\bf Problem 5} This is exactly the same problem
considered by Srivastava et al.~\cite{Srivastava14}. The goal
constraint is to be holding the red cylinder with an arbitrary
grasp. 39 blue cylinders crowd the table, blocking the red object.

\subsubsection*{\bf Problem 6} The goal constraints are that all
7 blue blocks must be on the left table and all 7 green
blocks must be on the right table. There are also 14 red blocks. The
close proximity of the blocks forces the planner to carefully order
its operations as well as to move red blocks out of the
way. 

\subsubsection*{\bf Solutions} 
The planner uses the same set of sampling
primitives on each problem with no problem-dependent information. Each
object has a possibly infinite number of grasps, computed from its
geometry. However, the first four generated are typically sufficient
to solve the problems posed here.  We restrict the PR2 to side grasps
to highlight the constrained nature of manipulation with densely
packed objects.  

We compared the performance of \algName{} when
$\proc{H} = \proc{H}_{\rm FF}$ to the performance of an unguided
version where $\proc{H}(s) = 0$ for all s.  We enforce a five-minute
timeout on all experiments.  There were 60 trials per problem all
conducted on a laptop with a 2.6GHz Intel Core i7 processor.  Each
entry in figure~\ref{table} reports the success percentage (\%) as
well as the median and median absolute deviation (MAD) of the runtime,
resulting plan length in terms of the number of actions, and number of
states visited in the forward search.  The median-based statistics are
used to be robust against outliers. 
Figure~\ref{figure:histogram} shows the distribution of runtimes for
$\proc{H}_{\rm FF}$ on problem 6. Several outliers cause the
mean runtime of 93 (the red line) to
be larger than the median runtime (the green line). The statistics for
trials that failed to find a solution are included in the
entries. Thus, entries with a runtime of 300 and MAD of 0 did not find
a plan for any trial. The accompanying video simulates the PR2 executing
a solution found by \algName{} for each problem.

Both versions of \algName{} solve problem 1
in several seconds despite the extremely high-dimensional
configuration space. A manipulation planning algorithm that explores
all adjacent modes~\cite{Hauser, HauserIJRR11} would be overwhelmed by
the branching factor despite the otherwise simple nature of the
problem.
The unguided search was unable to solve problems 4, 5, and 6
 in under 5 minutes due to the size of the
constructed state space and planning horizon.  
On problem 5, Srivastava et al. report a 63 percent success rate (over
randomly generated initial conditions) with an average runtime of 68
seconds. The median runtime of \algName{} is about a third of the
Srivastava et al. runtime with a 98 percent success rate.
Problem 6 is comparable in nature to but more complex than task 5
considered in FFRob~\cite{GarrettWAFR14}. Additionally, the runtime is
about half the FFRob reported median runtime of 157 seconds.

\subsubsection*{\bf Conclusions} 
The experiments show that by
leveraging the factored nature of common manipulation actions,
\algName{} is able to efficiently solve complex manipulation
tasks. The runtimes are improvements over runtimes on comparable
problems reported by Srivastava et al. and Garrett et
al.~\cite{GarrettWAFR14, Srivastava14}. Additionally, the dynamic
search allows \algName{} to solve regrasping, pushing, and stacking
problems all using the same planning algorithm. These results show
promise of \algName{} scaling to the large and diverse manipulation
problems prevalent in real world applications.

\addtolength{\textheight}{-10cm}   

\bibliographystyle{IEEEtran}
\bibliography{references}

\begin{thebibliography}{10}
\providecommand{\url}[1]{#1}
\csname url@rmstyle\endcsname
\providecommand{\newblock}{\relax}
\providecommand{\bibinfo}[2]{#2}
\providecommand\BIBentrySTDinterwordspacing{\spaceskip=0pt\relax}
\providecommand\BIBentryALTinterwordstretchfactor{4}
\providecommand\BIBentryALTinterwordspacing{\spaceskip=\fontdimen2\font plus
\BIBentryALTinterwordstretchfactor\fontdimen3\font minus
  \fontdimen4\font\relax}
\providecommand\BIBforeignlanguage[2]{{%
\expandafter\ifx\csname l@#1\endcsname\relax
\typeout{** WARNING: IEEEtran.bst: No hyphenation pattern has been}%
\typeout{** loaded for the language `#1'. Using the pattern for}%
\typeout{** the default language instead.}%
\else
\language=\csname l@#1\endcsname
\fi
#2}}

\bibitem{Simeon04}
T.~Sim\'eon, J.-P. Laumond, J.~Cort\'es, and A.~Sahbani, ``Manipulation
  planning with probabilistic roadmaps,'' \emph{IJRR}, vol.~23, pp. 729--746,
  2004.

\bibitem{Cambon}
S.~Cambon, R.~Alami, and F.~Gravot, ``A hybrid approach to intricate motion,
  manipulation and task planning,'' \emph{International Journal of Robotics
  Research}, vol.~28, 2009.

\bibitem{Hauser}
K.~Hauser and J.~Latombe, ``Integrating task and prm motion planning: Dealing
  with many infeasible motion planning queries,'' in \emph{ICAPS09 Workshop on
  Bridging the Gap between Task and Motion Planning}, 2009.

\bibitem{HauserIJRR11}
K.~Hauser and V.~Ng-Thow-Hing, ``Randomized multi-modal motion planning for a
  humanoid robot manipulation task,'' \emph{IJRR}, vol.~30, no.~6, pp.
  676--698, 2011.

\bibitem{Dogar12}
M.~R. Dogar and S.~S. Srinivasa, ``A planning framework for non-prehensile
  manipulation under clutter and uncertainty,'' \emph{Auton. Robots}, vol.~33,
  no.~3, pp. 217--236, 2012.

\bibitem{barry2013hierarchical}
J.~Barry, L.~P. Kaelbling, and T.~Lozano-Perez, ``A hierarchical approach to
  manipulation with diverse actions,'' in \emph{Robotics and Automation (ICRA),
  2013 IEEE International Conference on}.\hskip 1em plus 0.5em minus
  0.4em\relax IEEE, 2013, pp. 1799--1806.

\bibitem{Srivastava14}
S.~Srivastava, E.~Fang, L.~Riano, R.~Chitnis, S.~Russell, and P.~Abbeel,
  ``Combined task and motion planning through an extensible planner-independent
  interface layer,'' in \emph{IEEE Conference on Robotics and Automation
  (ICRA)}, 2014.

\bibitem{LozanoPerez81}
T.~Lozano-P\'erez, ``Automatic planning of manipulator transfer movements,''
  \emph{IEEE Transactions on Systems, Man, and Cybernetics}, vol.~11, pp.
  681--698, 1981.

\bibitem{handeyICRA87}
T.~Lozano-P\'{e}rez, J.~L. Jones, E.~Mazer, P.~A. O'Donnell, W.~E.~L. Grimson,
  P.~Tournassoud, and A.~Lanusse, ``Handey: A robot system that recognizes,
  plans, and manipulates,'' in \emph{IEEE International Conference on Robotics
  and Automation}, 1987.

\bibitem{Wilfong89}
G.~T. Wilfong, ``Motion planning in the presence of movable obstacles,'' in
  \emph{Symposium on Computational Geometry}, 1988, pp. 279--288.

\bibitem{Alami91}
R.~Alami, T.~Simeon, and J.-P. Laumond, ``A geometrical approach to planning
  manipulation tasks. the case of discrete placements and grasps,'' in
  \emph{ISRR}, 1990.

\bibitem{AlamiTwoProbs}
R.~Alami, J.-P. Laumond, and T.Sim\'eon, ``Two manipulation planning
  algorithms,'' in \emph{WAFR}, 1994.

\bibitem{dornhege09icaps}
C.~Dornhege, P.~Eyerich, T.~Keller, S.~Tr{\"u}g, M.~Brenner, and B.~Nebel,
  ``Semantic attachments for domain-independent planning systems,'' in
  \emph{International Conference on Automated Planning and Scheduling
  (ICAPS)}.\hskip 1em plus 0.5em minus 0.4em\relax AAAI Press, september 2009,
  pp. 114--121.

\bibitem{dornhege13irosws}
C.~Dornhege, A.~Hertle, and B.~Nebel, ``Lazy evaluation and subsumption caching
  for search-based integrated task and motion planning,'' in \emph{IROS
  workshop on AI-based robotics}, 2013.

\bibitem{Erdem}
E.~Erdem, K.~Haspalamutgil, C.~Palaz, V.~Patoglu, and T.~Uras, ``Combining
  high-level causal reasoning with low-level geometric reasoning and motion
  planning for robotic manipulation,'' in \emph{IEEE International Conference
  on Robotics and Automation (ICRA)}, 2011.

\bibitem{LagriffoulDSK12}
F.~Lagriffoul, D.~Dimitrov, A.~Saffiotti, and L.~Karlsson, ``Constraint
  propagation on interval bounds for dealing with geometric backtracking,'' in
  \emph{IEEE/RSJ International Conference on Intelligent Robots and Systems
  (IROS)}, 2012.

\bibitem{Plaku}
E.~Plaku and G.~Hager, ``Sampling-based motion planning with symbolic,
  geometric, and differential constraints,'' in \emph{IEEE International
  Conference on Robotics and Automation (ICRA)}, 2010.

\bibitem{HPN}
L.~P. Kaelbling and T.~Lozano-Perez, ``Hierarchical planning in the now,'' in
  \emph{IEEE Conference on Robotics and Automation (ICRA)}, 2011.

\bibitem{Pandey12}
A.~K. Pandey, J.-P. Saut, D.~Sidobre, and R.~Alami, ``Towards planning
  human-robot interactive manipulation tasks: Task dependent and human oriented
  autonomous selection of grasp and placement,'' in \emph{RAS/EMBS
  International Conference on Biomedical Robotics and Biomechatronics}, 2012.

\bibitem{GarrettWAFR14}
C.~R. Garrett, T.~Lozano-P\'{e}rez, and L.~P. Kaelbling, ``Ffrob: An efficient
  heuristic for task and motion planning,'' in \emph{International Workshop on
  the Algorithmic Foundations of Robotics (WAFR)}, 2014.

\bibitem{BonetG99}
B.~Bonet and H.~Geffner, ``Planning as heuristic search: New results,'' in
  \emph{Proc. of 5th European Conf. on Planning (ECP)}, 1999, pp. 360--372.

\bibitem{bonet2001planning}
------, ``Planning as heuristic search,'' \emph{Artificial Intelligence}, vol.
  129, no.~1, pp. 5--33, 2001.

\bibitem{HoffmannN01}
J.~Hoffmann and B.~Nebel, ``The {FF} planning system: Fast plan generation
  through heuristic search,'' \emph{Journal Artificial Intelligence Research
  (JAIR)}, vol.~14, pp. 253--302, 2001.

\bibitem{StilmanWAFR06}
M.~Stilman and J.~J. Kuffner, ``Planning among movable obstacles with
  artificial constraints,'' in \emph{Proceedings of the Workshop on Algorithmic
  Foundations of Robotics (WAFR)}, 2006.

\bibitem{openrave}
R.~Diankov and J.~Kuffner, ``Openrave: A planning architecture for autonomous
  robotics,'' Robotics Institute, Carnegie Mellon University, Tech. Rep.
  CMU-RI-TR-08-34, 2008.

\end{thebibliography}

\end{document}